\documentclass[times,twocolumn,final]{elsarticle}

\usepackage{medima}
\usepackage{framed,multirow}

\usepackage{amssymb}
\usepackage{latexsym}
\usepackage[T1]{fontenc}
\usepackage{url}
\usepackage{xcolor}
\usepackage{color}
\usepackage{amsmath,amssymb,amsfonts}
\usepackage{algorithmic}
\usepackage{graphicx}
\usepackage{textcomp}
\usepackage{booktabs}
\usepackage{bm}
\usepackage{amssymb}
\usepackage{url}
\usepackage{bbding}
\usepackage{array}
\definecolor{newcolor}{rgb}{.8,.349,.1}
\usepackage[colorlinks,linkcolor=red]{hyperref}
\journal{Medical Image Analysis}
\usepackage{geometry}
\begin{document}

\verso{Given-name Surname \textit{et~al.}}

\begin{frontmatter}

\title{Prompt Guiding Multi-Scale Adaptive Sparse Representation-driven Network for Low-Dose CT MAR}%
\author[1]{{Baoshun Shi}\corref{cor1}}
\cortext[cor1]{E-mail addresses: shibaoshun@ysu.edu.cn (B. Shi).}
\author[1,2]{Bing Chen}
\author[1,3]{Shaolei Zhang}
\author[4]{Huazhu Fu}
\author[5]{Zhanli Hu}

\address[1]{School of Information Science and Engineering, Yanshan University, Qinhuang Dao, 066004, China}
\address[2]{School of Electronics and Information Technology, Sun Yat-sen University, Guangzhou, 510275, China}
\address[3]{School of Computer Science and Technology, Beijing Institute of Technology, Beijing, 100081, China}
\address[4]{Institute of High Performance Computing, Agency for Science, Technology and Research, 138632, Singapore}
\address[5]{Institute of Lauterbur Research Center for Biomedical Imaging, Shenzhen Institute of Advanced Technology, Chinese Academy of Sciences, Shenzhen, 518055, China.}

\received{1 July 2024}
\finalform{10 August 2024}
\accepted{13 September 2024}
\availableonline{15 October 2024}
\communicated{S. Sarkar}

\begin{abstract}
Low-dose CT (LDCT) is capable of reducing X-ray radiation exposure, but it will potentially degrade image quality, even yields metal artifacts at the case of metallic implants. For simultaneous LDCT reconstruction and metal artifact reduction (LDMAR), existing deep learning-based efforts face two main limitations: $\emph{i}$) the network design neglects multi-scale and within-scale information; $\emph{ii}$) training a distinct model for each dose necessitates significant storage space for multiple doses. To fill these gaps, we propose a prompt guiding multi-scale adaptive sparse representation-driven network, abbreviated as PMSRNet, for LDMAR task. Specifically, we construct PMSRNet inspired from multi-scale sparsifying frames, and it can simultaneously employ within-scale characteristics and cross-scale complementarity owing to an elaborated prompt guiding scale-adaptive threshold generator (PSATG) and a built multi-scale coefficient fusion module (MSFuM). The PSATG can adaptively capture multiple contextual information to generate more faithful thresholds, achieved by fusing features from local, regional, and global levels. Furthermore, we elaborate a model interpretable dual domain LDMAR framework called PDuMSRNet, and train single model with a prompt guiding strategy for multiple dose levels. We build a prompt guiding module, whose input contains dose level, metal mask and input instance, to provide various guiding information, allowing a single model to accommodate various CT dose settings. Extensive experiments at various dose levels demonstrate that the proposed methods outperform the state-of-the-art LDMAR methods. Code is available in \href{https://github.com/shibaoshun/PDuMSRNet.git}{https://github.com/shibaoshun/PDuMSRNet.git}
\end{abstract}


\end{frontmatter}


\section{Introduction}
\label{sec:introduction}
X-ray computed tomography (CT) has been extensively utilized for medical diagnosis. However, the X-ray radiation increases the risk of cancer to patients when aggregated exposure surpasses the specific threshold (\citeauthor{9762023}, \citeyear{9762023}). Typically, a reduction in radiation dose is accomplished by managing the electrical current or voltage of the X-ray tube. Consequently, low-dose CT (LDCT) is extensively utilized to mitigate X-ray radiation risks. However, the decrease of the X-ray flux level leads to a sinogram (X-ray projection) corrupted by Poisson noise, subsequently yielding the reconstruction corrupted by noise. Additionally, when metallic implants are present within patients, the quality of LDCT images will be further reduced by metal artifacts (\citeauthor{9825711}, \citeyear{9825711}). Therefore, how to simultaneously perform low-dose CT denoising and metal artifact reduction (LDMAR) is a challenge.

With the continuous advancement of deep learning (DL) in medical image processing, various DL-based approaches using deep neural networks (DNNs) have been introduced into the LDCT denoising and full-dose CT (FDCT) MAR. Existing DL-based LDCT denoising or FDCT MAR methods can be primarily categorized into three classes: sinogram domain ( \citeauthor{8815915}, \citeyear{8815915}; \citeauthor{9439999}, \citeyear{9439999}), image domain (\citeauthor{8340157}, \citeyear{8340157}; \citeauthor{9320928}, \citeyear{9320928}; \citeauthor{8788607}, \citeyear{8788607}; \citeauthor{9362125}, \citeyear{9362125}; \citeauthor{9852485}, \citeyear{9852485};  \citeauthor{Wang2022DomainAD}, \citeyear{Wang2022DomainAD}; \citeauthor{Wang_2023}, \citeyear{Wang_2023}; \citeauthor{9928347}, \citeyear{9928347}; \citeauthor{mia2024ldct}, \citeyear{mia2024ldct}), and dual domain (  \citeauthor{10.1007/978-3-031-16446-0_71}, \citeyear{10.1007/978-3-031-16446-0_71}; \citeauthor{10.1007/978-3-030-87231-1_11}, \citeyear{10.1007/978-3-030-87231-1_11}; \citeauthor{10309982}, \citeyear{10309982}; \citeauthor{mia2025ldct1}, \citeyear{mia2025ldct1}; \citeauthor{mia2025ldct3}, \citeyear{mia2025ldct3}) methods. Specifically, sinogram domain methods are designed to correct the sinogram and subsequently apply the filtered back-projection (FBP) operation for CT image reconstruction. Image domain methods build DNNs to learn mapping between clean CT images and degraded CT images. In contrast, dual domain methods integrate knowledge from both the sinogram domain and the image domain to construct dual domain networks, achieving significant success. 
Although aforementioned methods have been developed for LDCT denoising and FDCT MAR tasks, applying previous techniques directly to the LDMAR task produces unsatisfactory reconstructions. To address this issue, \cite{9825711} first proposed DuDoUFNet, a dual domain under-to-fully-complete progressive restoration network for LDMAR. Nevertheless, DuDoUFNet neglects both multi-scale and within-scale information, limiting its performance, and it also suffers from inadequate model interpretability. In addition, the aforementioned methods suffer from large storage costs at the case of multiple dose levels since they often train each model for the specific setting. This ``one-model-for-one setting" approach not only increases storage overhead but also produces inflexible models tailored to single settings, limiting their adaptability in clinical applications. To overcome these bottlenecks, we propose a prompt guiding multi-scale adaptive sparse representation-driven network called PMSRNet which can train a single model for multiple doses, and incorporate it into a dual domain framework to build a model interpretable network called PDuMSRNet. The main contributions can be summarized as follows:
\begin{itemize}
	\item We come up with a prompt guiding multi-scale adaptive sparse representation-driven network dubbed as PMSRNet. It employs multi-scale sparsifying frames to establish a model-driven multi-scale adaptive network that incorporates both within-scale characteristics and cross-scale complementarity into the overall network. Specially, we leverage the dose map, metal mask and input instance as prompts, embedding them into a deep unfolded framework for solving the LDMAR task. PMSRNet can serve as a prior network embedded into deep unfolding networks \cite{10.1007/978-3-030-87231-1_11} for other image reconstruction tasks.
	\item We build a prompt guiding scale-adaptive threshold generator (PSATG) capable of adaptively extracting within-scale features for different scales to generate faithful thresholds during the shrinking procedure. The PSATG is composed of a shallow feature extraction module, a deep feature extraction module, a feature selection module, a threshold generating module, and a prompt guiding module. By using a prompt guiding module, the guiding information can be injected into the procedure of generating thresholds under the case of multiple dose levels.
	
	\item We propose a model interpretable dual domain LDMAR framework. Specifically, we formulate a sparsifying transform-based dual domain optimization model, and solve this optimization problem by using proximal gradient methods. Subsequently, we unfold this iteration algorithm into a deep neural network whose image domain subnetwork is the same architecture as that of PMSRNet, and the sinogram domain subnetwork consists of convolutional residual networks. The comprehensive experiments on the LDMAR task show that the proposed method can produce better results compared to SOTA methods.
	
\end{itemize}
\section{Related Work}
\subsection{LDCT Denoising} 
For LDCT denoising, numerous algorithms have been developed to improve the quality of LDCT images. Traditional methods can be summarized into three main groups, namely sinogram domain filtering (\citeauthor{6151161}, \citeyear{6151161}), iterative reconstruction (\citeauthor{8080249}, \citeyear{8080249}; \citeauthor{8081818}, \citeyear{8081818}), and image post-processing (\citeauthor{6610425}, \citeyear{6610425}). The sinogram domain filtering methods employ specific filters to directly suppress noise on raw projection data, potentially causing edge blurring and loss of structural information. The iterative reconstruction methods aim to minimize a unified objective function. However, these methods require raw projection data. In contrast, the image post-processing methods can directly process LDCT images, making them easily integrated into the CT machine system. 

Nonetheless, it is challenging for traditional hand-crafted priors to characterize all LDCT images. Recently, DL-based methods show promising reconstruction quality for LDCT denoising due to their powerful image distribution modeling capabilities. Some studies initially developed sinogram domain networks to correct sinograms, followed by using FBP to reconstruct CT images. \cite{5401721} adopted convolution residual networks to learn mapping between low-dose and full-dose sinograms. \cite{9439999} proposed an attention residual dense network address the low-dose sinogram denoising problem. Minor errors in sinograms may introduce new noise and artifacts into the reconstructed images. In contrast, image domain methods directly perform on LDCT images. \cite{7947200} proposed a residual encoder-decoder network to address the LDCT denoising issue. \cite{9320928} designed an densely connected convolutional neural network with a trainable Sobel convolution design. \cite{9362125} proposed a deep CNN denoising approach leveraging anatomical prior information and anatomical attention mechanisms for LDCT. \cite{Wang2022DomainAD} introduced a noise domain adaptive denoising network, which leveraged noise estimation and transfer learning. \cite{9852485} designed an unsupervised method with noise-characteristics modeling framework. \cite{Wang_2023} introduced a transformer-based network called CTformer, which incorporates dilated and cyclic shifts to comprehensively capture interactions. \cite{mia2024ldct} designed an unsupervised framework for LDCT task, demonstrating considerable potential for real-world CT imaging tasks lacking paired training data. Since noise stems from the back-projection of corrupted projection data, modeling the noise distribution is challenging. To address the limitation of single domain networks, dual domain methods leverage information from sinogram and image domains for LDCT denoising. \cite{8718010} introduced a dual domain network based on a 3D residual convolution network. \cite{10.1007/978-3-031-16446-0_71} designed a parallel network to fully and collaboratively utilized dual domain information. \cite{FU2023102787} proposed a generative adversarial network with attention encoding to achieve efficient full-dose reconstruction. \cite{mia2025ldct3} introduced a dual-domain joint optimization LDCT imaging framework for noisy sparse-view LDCT image reconstruction. \cite{mia2025ldct1} proposed a noise-inspired diffusion model for generalizable LDCT reconstruction, which requires only normal-dose data for training and can be effectively extended to various unseen dose levels. However, these LDCT algorithms do not account for the existence of metallic implants within patients. Therefore, directly applying them for LDMAR may lead to suboptimal reconstruction performance.
\subsection{FDCT MAR} 
For FDCT MAR, numerous methods have been proposed and have achieved outstanding performance. Physical effects correction methods model the physical effects of imaging and directly corrects metal-corrupted sinogram regions (\citeauthor{7268881}, \citeyear{7268881}). However, when high-atomic number metals are present, the metal trace region can be so severely damaged that the above correction methods fail to produce satisfactory results (\citeauthor{8331163}, \citeyear{8331163}). Sinogram completion methods utilize various estimation techniques to fill the metal trace region within sinogram (\citeauthor{5401722}, \citeyear{5401722}). As it is difficult to generate perfect estimation, these estimation methods often introduce secondary artifacts. To cope with this problem, iterative reconstruction methods (\citeauthor{8576532}, \citeyear{8576532}; \citeauthor{6522894}, \citeyear{6522894}) are used to reconstruct high-quality CT images from the observed data, but these methods require suitable hand-crafted regularization terms.

With the widespread application of DNNs in medical image processing, numerous DL-based methods have been developed for MAR. \cite{8815915} employed generative adversarial network to correct metal-corrupted sinogram. Since sinogram domain methods are difficult to produce perfect estimation of the metal trace region, secondary artifacts will inevitably be introduced to the reconstructed image. Image domain methods regard the MAR task as image restoration. \cite{8331163} utilized DNNs to generate the prior image that assists in correcting the metal-corrupted regions of the sinogram. \cite{8788607} introduced an unsupervised artifact disentanglement network that produces artifact-reduced images using unpaired training data. The primary limitation of image domain methods is the neglect of the consistency constraint. To address the issue of single domain networks, dual domain methods that exploit dual domain knowledge have been developed these years. The seminal work termed as DuDoNet was proposed by \cite{8953298}, and a Radon inversion layer was designed to bridge between the sinogram domain and the image domain network. \cite{9765584} proposed a parallel dual domain network by modifying the fixed priority mode of the previous dual domain network. \cite{10.1007/978-3-030-87231-1_11} proposed a model-driven dual domain network, integrating imaging geometrical model constraints into the mutual learning process of sinogram and image domain information. \cite{WANG2023102729} proposed a deep unfolding dual domain network by unfolding the iterative algorithm for MAR task. Although the dual domain network mentioned above have achieved high-quality CT images, the lack of domain knowledge during training remains a significant limitation in improving the MAR performance. \cite{Shi2024MudNetMD} proposed a method that leverages sinogram domain, image domain, artifact domain, and coding domain knowledge to construct a model-driven network for MAR. \cite{Shi2024CouplingMA} proposed a novel dual domain network that couples model- and data-driven networks for MAR. Although these networks have achieved excellent performance in MAR tasks, they are designed for full-dose CT and do not address the LDMAR issue. Furthermore, the design of these networks has not considered the incorporation of multi-scale and within-scale information. This leaves room for further improvement. 

\section{The proposed PDuMSRNet for Dual Domain LDMAR Network}
\begin{figure*}[!t]
	\centering
	{\includegraphics[width=1.0\linewidth]{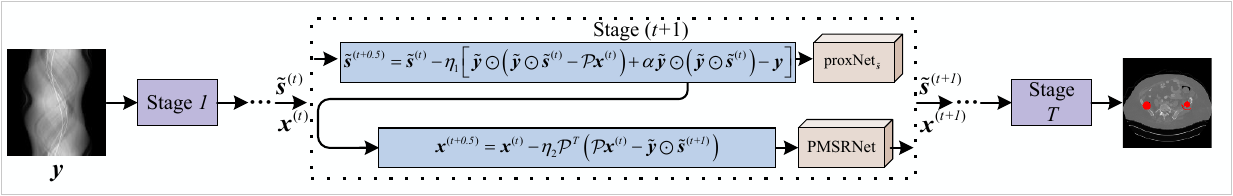}}
	\caption{Overview of our PDuMSRNet. The architecture, which is unfolded from the optimization iterations for solving the LDMAR task, comprises $T$ stages corresponding to $T$ iterative steps. Specifically, the sinogram domain variable $\tilde{\textbf{\emph{s}}}^{(t+1)}$ and the image domain variable $\textbf{\emph{x}}^{(t+1)}$ are updated alternately using the proxNet$_{{\tilde{\textbf{\emph{s}}}}}$ for sinogram-domain processing and PMSRNet for image-domain reconstruction at the ($t+1$)-th stage. The architecture of proxNet${\tilde{\textbf{\emph{s}}}}$ comprises three residual blocks in sinogram domain, whereas PMSRNet is the proposed prompt guiding multi-scale adaptive sparse representation-driven network in image domain. This design explicitly follows the update rules derived in Eqn. (\ref{eq:9}) and Eqn. (\ref{eq:14}), ensuring a model-driven deep learning framework.}
	\label{Fig:1}
\end{figure*}
\subsection{Model Design and Optimization}
Assume that $\textbf{\emph{x}}$ represents the clean CT image, the measurement $\textbf{\emph{y}}$ can be modeled as
\begin{equation}
	\textbf{\emph{y}} = \mathcal{P}\textbf{\emph{x}} + \textbf{\emph{n}}
	\label{eq:1}
\end{equation}
where $\mathcal{P}$ denotes the forward projection (FP) operation under the LDCT condition, and $\textbf{\emph{n}}$ denotes the noise. The corresponding inverse problem for this linear system in Eqn. (\ref{eq:1}) is ill-posed, and specific regularization is necessary to address solution ambiguity and suppress noise magnification. A typical optimization model can be expressed as
\begin{equation}
	\mathop {\min }\limits_{\textbf{\emph{x}}} ||\mathcal{P}\textbf{\emph{x}} - \textbf{\emph{y}} || _2^2+\lambda\emph{R}( \textbf{\emph{x}})
	\label{eq:2}
\end{equation}
where $\lambda$ is a regularization parameter, and $R(\cdot)$ serves as a regularizer to provide prior information about $\textbf{\emph{x}}$. To facilitate the enhancement of dual domain information and joint regularization, the optimization model can be formulated
\begin{equation}
	\mathop {\min }\limits_{\textbf{\emph{s}}, \textbf{\emph{x}}} ||\mathcal{P}\textbf{\emph{x}} - \textbf{\emph{s}}|| _2^2+\alpha|| \textbf{\emph{s}} - \textbf{\emph{y}} || _2^2 + \lambda_1 \emph{R}_1( \textbf{\emph{s}})+\lambda_2\emph{R}_2(\textbf{\emph{x}}) 
	\label{eq:3}
\end{equation}
where $\alpha$ is a weight parameter used to balance different data fidelity terms, and $\textbf{\emph{s}}$ is the clean sinogram. In Eqn. (\ref{eq:3}), $\lambda_1$ and $\lambda_2$ are regularization parameters. $R_1(\cdot)$ and $R_2(\cdot)$ denote the regularization terms that enforce certain desirable properties onto $\textbf{\emph{s}}$ and $\textbf{\emph{x}}$, respectively. In fact, sinograms with more uniform profiles are easier to restore (\citeauthor{10.1007/978-3-030-87231-1_11}, \citeyear{10.1007/978-3-030-87231-1_11}). Therefore, we rewrite $\textbf{\emph{s}}$ as
\begin{equation}
	\textbf{\emph{s}} = \tilde{\textbf{\emph{y}}} \odot \tilde{\textbf{\emph{s}}}
	\label{eq:4}
\end{equation}
where $\odot$ represents the element-wise product, and $\tilde{\textbf{\emph{y}}}$ represents the normalization coefficient, which are typically obtained through a U-Net architecture (\citeauthor{10.1007/978-3-030-87231-1_11}, \citeyear{10.1007/978-3-030-87231-1_11}). By substituting Eqn. (\ref{eq:4}), the dual domain LDMAR problem in Eqn. (\ref{eq:3}) can be rewritten as
\begin{align}
	\mathop {\min }\limits_{\tilde{\textbf{\emph{s}}}, \textbf{\emph{x}}} &|| \mathcal{P}\textbf{\emph{x}} - \tilde{\textbf{\emph{y}}} \odot \tilde{\textbf{\emph{s}}} || _2^2+\alpha|| \tilde{\textbf{\emph{y}}} \odot \tilde{\textbf{\emph{s}}} - \textbf{\emph{y}} || _2^2+\lambda_1\emph{R}_1( \tilde{\textbf{\emph{s}}}) +\lambda_2\emph{R}_2(\textbf{\emph{x}}).
	\label{eq:5}
\end{align}
Additionally, we utilize a learnable sparsifying frame to explore the
structural information of CT images (\citeauthor{10198706}, \citeyear{10198706}; \citeauthor{10034942}, \citeyear{10034942}; \citeauthor{9763013}, \citeyear{9763013}). The optimization model can be further expressed as
\begin{align}
	\mathop {\min }\limits_{\tilde{\textbf{\emph{s}}}, \textbf{\emph{x}}} & || \mathcal{P}\textbf{\emph{x}} - \tilde{\textbf{\emph{y}}} \odot \tilde{\textbf{\emph{s}}} || _2^2+\alpha|| \tilde{\textbf{\emph{y}}} \odot \tilde{\textbf{\emph{s}}} - \textbf{\emph{y}} || _2^2+\lambda_1\emph{R}_1(\tilde{\textbf{\emph{s}}}) +\lambda_2|| {\textbf{\emph{W}}}\textbf{\emph{x}}||_1
	\label{eq:6}
\end{align}
where $|| \cdot||_1$ represents the $\ell_1$ norm. To address the optimization problem defined in Eqn. (\ref{eq:6}), we alternately solve $\tilde{\textbf{\emph{s}}}$ and $\textbf{\emph{x}}$. At the ($t+1$)-th iteration, $\tilde{\textbf{\emph{s}}}$ and $\textbf{\emph{x}}$ are solved alternately as follows.\par
Updating $\tilde{\textbf{\emph{s}}}$: The iteration for solving $\tilde{\textbf{\emph{s}}}$ can be derived as
\begin{align}
	\tilde{\textbf{\emph{s}}}^{(t+1)}=&\mathop {\arg\min}\limits_{\tilde{\textbf{\emph{s}}}} \Big\{||\mathcal{P}\textbf{\emph{x}}^{(t)} - \tilde{\textbf{\emph{y}}} \odot \tilde{\textbf{\emph{s}}}||_2^2+\alpha|| \tilde{\textbf{\emph{y}}} \odot \tilde{\textbf{\emph{s}}} - \textbf{\emph{y}} ||_2^2 +\lambda_1\emph{R}_1(\tilde{\textbf{\emph{s}}})\Big\}.
	\label{eq:7}
\end{align}
The quadratic approximation with respect to $\tilde{\textbf{\emph{s}}}$ in Eqn. (\ref{eq:7}) can be written as
\begin{align}
	\tilde{\textbf{\emph{s}}}^{(t+1)}=&\mathop {\arg\min}\limits_{\tilde{\textbf{\emph{s}}}} \Big\{\frac{1}{2}||\tilde{\textbf{\emph{s}}}-[ \tilde{\textbf{\emph{s}}}^{(t)}-\eta_1\bigtriangledown\emph{f}_1( \tilde{\textbf{\emph{s}}}^{(t)})] ||_2^2 +\lambda_1\eta_1\emph{R}_1(\tilde{\textbf{\emph{s}}})\Big\}
	\label{eq:8}
\end{align}
where $\emph{f}_1(\tilde{\textbf{\emph{s}}}^{(t)})= || \mathcal{P}\textbf{\emph{x}}^{(t)} -\tilde{\textbf{\emph{y}}} \odot \tilde{\textbf{\emph{s}}}^{(t)}|| _2^2 + \alpha|| \tilde{\textbf{\emph{y}}} \odot \tilde{\textbf{\emph{s}}}^{(t)} - \textbf{\emph{y}} || _2^2$, and $\eta_1$ is the stepsize. Then, the updating rule of $\tilde{\textbf{\emph{s}}}$ can be deduced as
\begin{equation}
	\tilde{\textbf{\emph{s}}}^{(t+1)}={\rm prox}_{\lambda_1\eta_1}( \tilde{\textbf{\emph{s}}}^{(t+0.5)})
	\label{eq:9}
\end{equation}
where ${\rm prox}_{\lambda_1\eta_1}(\cdot)$ is a proximal operator linked to the regularization term $\emph{R}_1(\cdot)$ about $\tilde{\textbf{\emph{s}}}$. Based on $\tilde{\textbf{\emph{s}}}^{(t+0.5)}=\tilde{\textbf{\emph{s}}}^{(t)}-\eta_1\bigtriangledown\emph{f}_1( \tilde{\textbf{\emph{s}}}^{(t)})$ and $\bigtriangledown\emph{f}_1 ( \tilde{\textbf{\emph{s}}}^{(t)})=\tilde{\textbf{\emph{y}}}\odot(\tilde{\textbf{\emph{y}}}\odot\tilde{\textbf{\emph{s}}}^{(t)}-\mathcal{P}\textbf{\emph{x}}^{(t)})+\alpha[\tilde{\textbf{\emph{y}}}\odot(\tilde{\textbf{\emph{y}}}\odot\tilde{\textbf{\emph{s}}}^{(t)}-\textbf{\emph{y}})]$, we have 

\begin{align}
	\tilde{\textbf{\emph{s}}}^{(t+0.5)}=&\tilde{\textbf{\emph{s}}}^{(t)}-\eta_1[\tilde{\textbf{\emph{y}}}\odot(\tilde{\textbf{\emph{y}}}\odot\tilde{\textbf{\emph{s}}}^{(t)}-\mathcal{P}\textbf{\emph{x}}^{(t)}) \nonumber \\
	&+\alpha\tilde{\textbf{\emph{y}}}\odot(\tilde{\textbf{\emph{y}}}\odot\tilde{\textbf{\emph{s}}}^{(t)})-\textbf{\emph{y}}].
	\label{eq:10}
\end{align}

Updating $\textbf{\emph{x}}$:
Similarly, the iteration for solving $\textbf{\emph{x}}$ can be derived as
\begin{equation}
	\textbf{\emph{x}}^{(t+1)}=\mathop {\arg\min}\limits_\textbf{\emph{x}}\Big\{||\mathcal{P}\textbf{\emph{x}} - \tilde{\textbf{\emph{y}}} \odot \tilde{\textbf{\emph{s}}}||_2^2 +\lambda_2||{\textbf{\emph{W}}}\textbf{\emph{x}}||_1\Big\}.
	\label{eq:11}
\end{equation}
The quadratic approximation with respect to $\textbf{\emph{x}}$ in Eqn. (\ref{eq:11}) can be written as
\begin{align}
	\textbf{\emph{x}}^{(t+1)}=&\mathop {\arg\min}\limits_\textbf{\emph{x}}\Big\{\frac{1}{2}|| \textbf{\emph{x}}-[ \textbf{\emph{x}}^{(t)}-\eta_1\bigtriangledown\emph{f}_2( \textbf{\emph{x}}^{(t)}) ] || _2^2 \nonumber\\
	&+\lambda_2 \eta_2|| {\textbf{\emph{W}}}\textbf{\emph{x}}||_1\Big\}
	\label{eq:12}
\end{align}
where $\bigtriangledown \emph{f}_2(\textbf{\emph{x}}^{(t)})= \mathcal{P}^{T}(\mathcal{P}\textbf{\emph{x}}^{(t)} - \tilde{\textbf{\emph{y}}} \odot \tilde{\textbf{\emph{s}}}^{(t+1)})$, and $\eta_2$ is the stepsize. 
The updating rule of $\textbf{\emph{x}}$ can be written as
\begin{equation}
	\textbf{\emph{x}}^{(t+1)}={\rm prox}_{\lambda_2\eta_2}(\textbf{\emph{x}}^{(t+0.5)})
	\label{eq:13}
\end{equation}
where $\textbf{\emph{x}}^{(t+0.5)} = \textbf{\emph{x}}^{(t)} - \eta_2\mathcal{P}^{T}(\mathcal{P}\textbf{\emph{x}}^{(t)} - \tilde{\textbf{\emph{y}}} \odot \tilde{\textbf{\emph{s}}}^{(t+1)})$. Eqn. (\ref{eq:13}) can be approximately solved as
\begin{equation}
	\textbf{\emph{x}}^{(t+1)}={\textbf{\emph{W}}}^\mathrm{T}S_{\varepsilon} ({\textbf{\emph{W}}} \textbf{\emph{x}}^{(t+0.5)})
	\label{eq:14}
\end{equation}
where ${\textbf{\emph{W}}}$ represents the trainable sparsifying frame, and $S_{\varepsilon}(\cdot)$ is the soft threshold function defined as $soft_{\varepsilon}(\mu)=sign(\mu) max(|\mu| - \varepsilon, 0)$.
\subsection{Dual Domain Model Interpretable Network}
While existing deep learning-based LDMAR algorithms \cite{9825711} are data-driven, they lack model interpretability, which is crucial for building models based on existing knowledge and facilitating subsequent theoretical analysis. Constructing an interpretable LDMAR network therefore remains a challenge. To address this challenge, we leverage deep unfolding techniques, which have shown remarkable success in computer vision tasks. Each iterative step in Eqn. (\ref{eq:9}) and Eqn. (\ref{eq:14}) is transformed into a corresponding network module, thus constructing dual-domain LDMAR networks. The resulting network incorporating the prompt guiding module is referred to as PDuMSRNet, while the version without it is termed DuMSRNet. Since each module in PDuMSRNet directly corresponds to an operator in the iterative algorithm, the proposed architecture possesses inherent model interpretability.

As shown in Fig.~\ref{Fig:1}, PDuMSRNet is composed of $T$ stages, corresponding to $T$ iterations of the unfolding operations. At each stage, PDuMSRNet consists of a $\tilde{\textbf{\emph{s}}}$-Net and a $\textbf{\emph{x}}$-Net. According to the update rules defined in Eqn. (\ref{eq:9}) and Eqn. (\ref{eq:14}), we incrementally construct $\tilde{\textbf{\emph{s}}}$-Net and $\textbf{\emph{x}}$-Net.

\textbf{$\tilde{\textbf{\emph{s}}}$-Net:} At the ($t$+1) stage, given $\tilde{\textbf{\emph{s}}}^{(t)}$, $\textbf{\emph{x}}^{(t)}$, and $\tilde{\textbf{\emph{y}}}$, $\tilde{\textbf{\emph{s}}}^{(t+0.5)}$ is computed and passed to proxNet$_{\tilde{\textbf{\emph{s}}}}(\cdot)$ to execute the operator prox$_{\lambda_1\eta_1}(\cdot)$. The proxNet$_{\tilde{\textbf{\emph{s}}}}(\cdot)$ consists of several convolutional residual blocks. 

\textbf{$\textbf{\emph{x}}$-Net:} Similarly, given $\tilde{\textbf{\emph{s}}}^{(t+1)}$, $\textbf{\emph{x}}^{(t)}$, and $\tilde{\textbf{\emph{y}}}$, the CT image can be updated by the proposed PMSRNet. 
\begin{figure*}[!t]
	\centering
	{\includegraphics[width=1.0\linewidth]{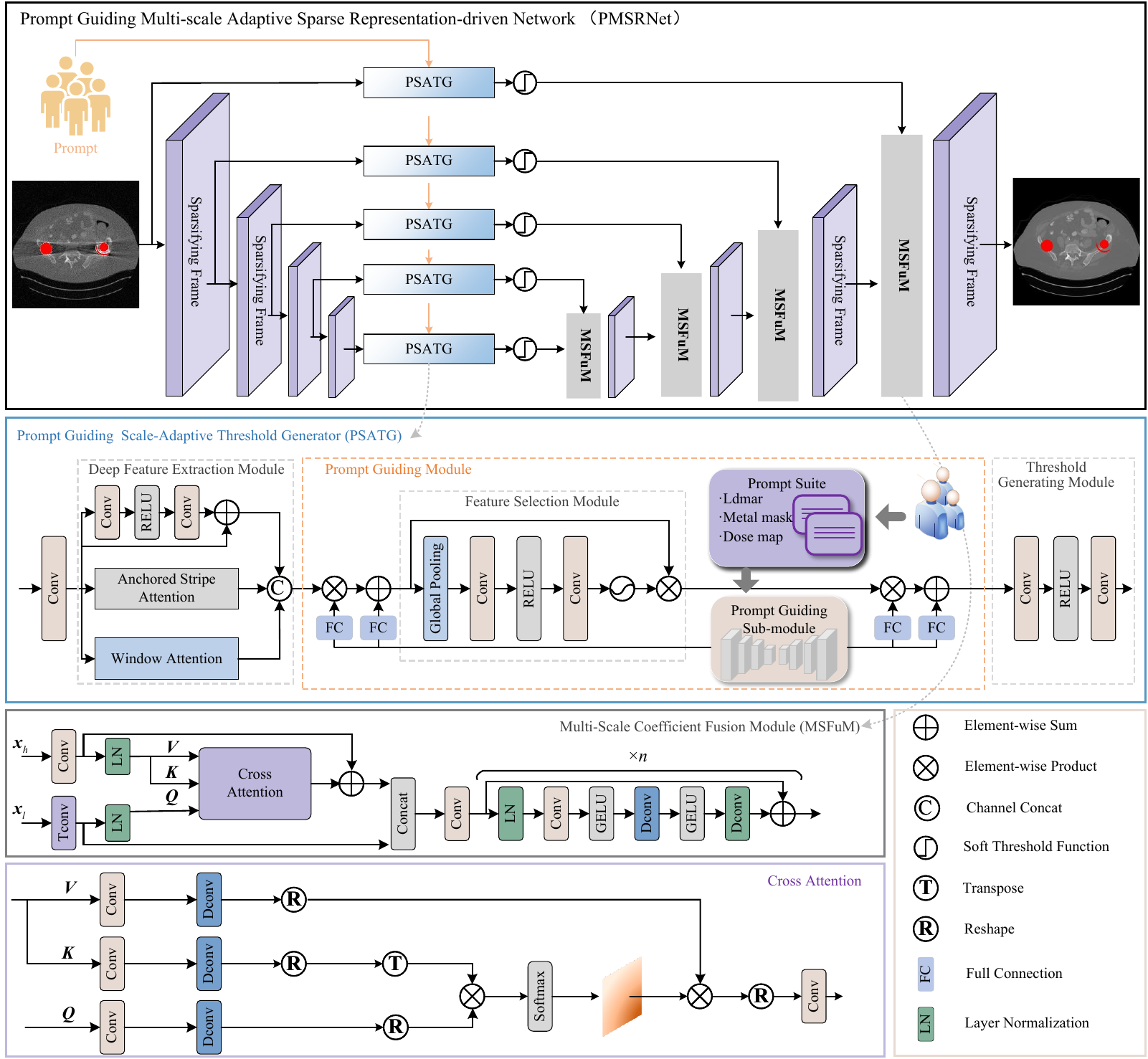}}
	\caption{{Overview of our prompt guiding multi-scale adaptive sparse representation-driven network named PMSRNet (top), our prompt guiding scale-adaptive threshold generator termed as PSATG (middle) and multi-scale coefficient fusion module called MSFuM (bottom). The PMSRNet contains three components: multi-scale sparsifying frames, PSATG and MSFuM. The PSATG is designed to generate faithful thresholds with the prior of prompt information and contains five parts: 1) The shallow feature extraction module relies on a convolution layer; 2) The deep feature extraction module models images at local, regional, and global levels by parallelizing convolutional neural networks, window attention, and anchored stripe attention; 3) The feature selection module utilizes channel attention mechanism to adaptively reweight information from different feature extraction module; 4) The threshold generating module enhances the performance of PMSRNet by generating faithful thresholds using information from the feature selection module; 5) The prompt guiding module utilizes prompt information, including LDCT images with metallic implants, metal masks, and dose maps, to guide the threshold generating process. The MSFuM which is used to fuse the multi-scale representations mainly contains a cross attention and a feed-forward network to effectively integrate features from different resolutions. Structurally, the PMSRNet adopts a four-scale architecture. At each scale, the processing flow consists of a sparse coding procedure by learnable sparsifying frame, a threshold processing procedure with the prior of prompt information, and an image generation procedure by learnable sparsifying frame, which work in sequence to ultimately produce the final reconstructed image.}}
	\label{Fig:2}
\end{figure*}

\section{Prompt Guiding Multi-Scale Adaptive Sparse Representation-Driven Network}
\label{sec3:Prompt Guiding Multi-Scale Adaptive Sparse Representation-Driven Network}
\subsection{Architecture Principles}
Multi-scale features and cross-scale complementarity are important for solving image inverse problems. \cite{NEURIPS2022_5b288823} proposed MSANet that effectively integrates the within-scale characteristics and cross-scale complementarity for image denoising. Inspired by this, we propose a prompt guiding multi-scale sparse representation-driven network for LDMAR. In PMSRNet, we employ a learnable sparsifying frame to filter the input data for noise removal. More precisely, this process can be expressed as
\begin{equation}
	\bar{\textbf{{\emph{x}}}}={\textbf{\emph{W}}}^\mathrm{T}S_{\varepsilon} ({\textbf{\emph{W}}} \tilde{\textbf{\emph{x}}})
	\label{eq:15}
\end{equation}
where $\tilde{\textbf{\emph{x}}}$ represents the input LDCT image with metallic implants, $\bar{\textbf{{\emph{x}}}}$ represents the filtered CT image. We extend the single-scale sparsifying frame to a multi-scale version to enhance its representational capacity. The multi-scale sparsifying frames are sequential structure where information flows between different scales while remaining unchanged within each scale. To obtain multi-scale features, we set the filter sizes in the sparsifying frame network to 5$\times$5, 7$\times$7, 9$\times$9, and 11$\times$11, with corresponding output channel numbers of 25, 49, 81, and 121, respectively. Given the critical role of the threshold in the soft threshold function, we develop a prompt guiding scale-adaptive threshold generator to generate faithful thresholds.

As illustrated in Fig.~\ref{Fig:2}, PMSRNet adopts a four-scale architecture consisting of sparse coding procedure, threshold processing procedure, and generating procedure. Specifically, as the number of layers increases, the sparse coding procedure reduces the image resolution by half while also increasing the number of channels to extract features at four different scales. For the threshold processing procedure, the prompt guiding scale-adaptive threshold generator (PSATG) models the features extracted from the sparse coding procedure at local, regional, and global levels to utilize the within-scale characteristics. To attain the desired cross-scale complementarity, the generating procedure involves the fusion of features from high-resolution frame coefficients and low-resolution frame coefficients through a multi-scale coefficient fusion module. In addition, with the decrease in the number of layers, the generating procedure gradually increases the feature resolution while decreasing the number of channels. Next, we introduce the proposed PSATG and MSFuM in detail.

\subsection{Prompt Guiding Scale-Adaptive Threshold Generator}
According to optimization theory, spatially variant thresholds are essential for effective image denoising in the deep sparsifying transform model, as noise levels vary with radiation dose level and artifact characteristics depend on metal implants. Thus, we employ a prompt learning strategy. Recently popularized for all-in-one IR tasks (\citeauthor{promptir}, \citeyear{promptir}), prompt learning generalizes across degradation types and levels and achieves SOTA results in multi-task IR frameworks. In LDMAR, prompts derived from dose levels and metal implants generate adaptive thresholds, addressing both low-dose CT tasks under varying dose levels and MAR task for diverse metal implants.\par	
Contextual information is crucial for both high- and low-resolution layers. High-resolution features lose details when downsampled, while low-resolution features compromise structural consistency. To overcome these issues and leverage cross-scale complementarity, we propose a prompt guiding scale-adaptive threshold generator (PSATG) that fuses prompt features with multi-scale features from local, regional, and global levels. PSATG comprises shallow and deep feature extraction modules, a feature selection module, a threshold generation module, and a prompt guiding module. The shallow feature extraction module uses a convolutional layer. The deep feature extraction module includes branches for local (LIEB), regional (RIEB), and global (GIEB) information extraction. LIEB employs a convolutional residual structure for local information, while RIEB and GIEB utilize window attention (\citeauthor{9879380}, \citeyear{9879380}) and anchored stripe attention (\citeauthor{8953616}, \citeyear{8953616}) for regional and global information, respectively. Feature maps from the three branches are processed independently and concatenated along the channel dimension. The prompt guiding module directs threshold generation using prompt information. In processing flow of PSATG, the shallow feature extraction module first processes the input to produce ${\hat{\bm x}}$. The three branches then extract local, regional, and global features from ${\hat{\bm x}}$. Finally, the feature selection module reweights channels from these branches to select information more conducive to LDCT image reconstruction.\par
The prompt guiding module employs a prompt suite comprising LDMAR images, metal masks, and dose maps to guide the feature extraction process. Specifically, the LDCT image with metallic implants is the input of DuMSRNet. The dose map is defined as a matrix in which each element is the reciprocal of the corresponding dose level value. The metal mask is obtained through a simple thresholding segmentation of LDCT image with metallic implants. These three prompts are integrated via channel concatenation. The concatenated features are then passed to two separate fully connected layers to generate a weight vector and a bias term. An element-wise multiplication of the weights with the features is performed, followed by the addition of the bias. The resulting reweighted features are subsequently fed into the threshold generation module. Notably, the PSATG employs shared parameters across all unfolding stages during training, which enhances both model efficiency and cross-stage consistency.\par 
As illustrated in Fig.~\ref{Fig:2}, we take the LDCT images with metallic implants, the metal masks and the dose maps as prompts, and perform a channel-wise concatenation operation on them to obtain the three-channel image $\textbf{P}$, i.e.,\par
\begin{equation}
	\textbf{P} ={\rm Concat}(\tilde{\textbf{\emph{x}}}, \bm m, \bm d),
	\label{eq:16}
\end{equation}
where $\bm m$ and $\bm d$ denote the metal mask and dose map, respectively. \par 
A prompt guiding sub-module represented by ${F_{ext}({\bullet})}$ is employed to extract features from the prompt $\textbf{P}$. Subsequently, these features are processed through fully connected layers ${\rm FC({\bullet})}$ to achieve the appropriate shape (1$\times$channel or dimension) required for the corresponding module's outputs. Finally, the output features are multiplied by weight $\textbf{\emph{k}}$ and added by bias $\textbf{\emph{b}}$. Mathematically, the process of PSATG can be formulated as
\begin{equation}
	\textbf{\emph{k}}_{1} = {\rm FC}_{1}(F_{ext}(\textbf{P})),\textbf{\emph{k}}_{2} = {\rm FC}_{2}(F_{ext}(\textbf{P})),
	\label{eq:17}
\end{equation}

\begin{equation}
	\textbf{\emph{b}}_{1} = {\rm FC}_{3}(F_{ext}(\textbf{P})),\textbf{\emph{b}}_{2} = {\rm FC}_{4}(F_{ext}(\textbf{P})),
	\label{eq:18}
\end{equation}

\begin{equation}
	\textbf{\emph{x}}_{in} = {\rm FSM}(\textbf{\emph{k}}_{1}[{\rm LIEB}(\textbf{\emph{x}}_{in}),{\rm RIEB}(\textbf{\emph{x}}_{in}),{\rm GIEB}(\textbf{\emph{x}}_{in})]+\textbf{\emph{b}}_{1}),
	\label{eq:19}
\end{equation}
and
\begin{equation}
	\textbf{\emph{x}}_{out} = {\rm Conv}({\rm ReLU}( {\rm Conv}({\textbf{\emph{k}}_{2}}\textbf{\emph{x}}_{in}+{\textbf{\emph{b}}_{2}})))
	\label{eq:20}
\end{equation}
where FSM($\cdot$) denotes the feature selection module, and [$\textbf{\emph{x}}_{1}$, $\textbf{\emph{x}}_{2}$, $\textbf{\emph{x}}_{3}$] denotes the channel-wise concatenation of $\textbf{\emph{x}}_{1}$, $\textbf{\emph{x}}_{2}$, and $\textbf{\emph{x}}_{3}$.

To obtain global range information, we utilize the anchored stripe attention proposed in \cite{8953616} to model image global range hierarchies. Specifically, we replace the direct comparison of queries and keys with anchors as intermediaries for similarity comparison. Then, we alternate between horizontal and vertical stripes to balance the modeling capacity for global range and the computational complexity. 

\subsection{Multi-Scale Coefficient Fusion Module}
High-resolution features possess abundant image details, while low-resolution features contain a wealth of contextual information. To fully leverage cross-scale complementarity, we construct a multi-scale coefficient fusion module (MSFuM) based on the cross attention mechanism, effectively fusing information from multiple scales. The MSFuM within PMSRNet combines the detailed information from high-resolution features with the contextual information from low-resolution features to obtain complementary multi-scale features across different scales. By doing so, PMSRNet can embrace the complementarity of multi-scale and within-scale information, thereby significantly enhancing the representation ability. As shown in Fig.~\ref{Fig:2}, MSFuM establishes a double-cross attention to fuse high-resolution and low-resolution features. Specifically, we first upsample the low-resolution features $\textbf{\emph{x}}_l$ to match the resolution of the high-resolution features $\textbf{\emph{x}}_h$, while processing $\textbf{\emph{x}}_h$ with a convolutional layer. Then, the transformed $\textbf{\emph{x}}_l$ and $\textbf{\emph{x}}_h$ are normalized by the LayerNorm. The input $\textbf{\emph{Q}}$ of cross attention originates from $\textbf{\emph{x}}_l$, while $\textbf{\emph{K}}$ and $\textbf{\emph{V}}$ are obtained from $\textbf{\emph{x}}_h$. Next, we embed them through convolution and utilize deep convolution to encode channel-wise contextual information. Finally, a reshape operation is utilized to reformulate $\textbf{\emph{Q}}$, $\textbf{\emph{K}}$, and $\textbf{\emph{V}}$ into tokens. Mathematically, this process can be defined as
\begin{align}
	\left\{
	\begin{aligned}
		\tilde{\textbf{\emph{Q}}}&=\rm R(Dconv(Conv(\textbf{\emph{Q}}))\\
		\tilde{\textbf{\emph{K}}}&=\rm R(Dconv(Conv(\textbf{\emph{K}}))\\
		\tilde{\textbf{\emph{V}}}&=\rm R(Dconv(Conv(\textbf{\emph{V}}))
	\end{aligned}
	\right.
	\label{eq:21}
\end{align}
where Dconv denotes the depth-wise convolution, and $R(\cdot)$ denotes the reshape operation. The softmax function is applied to generate the attention map, which can be expressed as
\begin{equation}
	\textbf{\emph{C}}=\rm SoftMax(\tilde{\textbf{\emph{K}}}^T\tilde{\textbf{\emph{Q}}}).
	\label{eq:22}
\end{equation}
Here, $\rm{SoftMax(} \bullet {)}$ is the softmax function applied to re-weight the matrix multiplication. Next, we employ the reshape and convolution operations to further extract features. In summary, the cross attention of MSFuM can be expressed as
\begin{equation}
	\rm CAttention( \textbf{\emph{Q}}, \textbf{\emph{K}}, \textbf{\emph{V}}) = \rm Conv(R(\tilde{\textbf{\emph{V}}}\textbf{\emph{C}})).
	\label{eq:23}
\end{equation}
The proposed MSFuM is performed between high-resolution and low-resolution features to extensively explore their correlation, enhancing crucial information and mutually compensating for reconstruction deficiencies.
\section{Experiments Setting and Results}
\subsection{Loss Function}
During training, we adopts a hybrid dual domain loss function to achieve optimal reconstruction results. First, we adopt mean squared error (MSE) loss function, which can be defined as
\begin{equation}
	\mathcal L_{mse} = ||{\bm m}\odot({\bm {x}_{out}}-{\bm {x}})||_2^2 + \omega_1 ||({\bm {s}_{out} - \bm {s}})||_2^2
	\label{eq:24}
\end{equation}
where ${\bm m}$, ${\bm {x}}$ and ${\bm {s}}$ represent a binary non-metal mask, the ground truth FDCT image without metallic implants and ground truth full-dose sinogram without metallic implants, respectively. In Eqn. (\ref{eq:24}), $\omega_1$ is a weight parameter used to balance different loss components.
However, the smooth nature of MSE makes it susceptible to outliers, resulting in the creation of new artifacts. To overcome this problem, consistent with \cite{9320928} and \cite{9928347}, we adopt the multi-scale perceptual (MSP) loss using a pre-trained ResNet-50 (\citeauthor{7780459}, \citeyear{7780459}), which can be defined as
\begin{equation}
	\mathcal L_{msp} = \sum_{n=1}^{N}||\Psi_n({\bm m}\odot{\bm {x}_{out}})-\Psi_n({\bm m}\odot{\bm {x}})||_2^2
	\label{eq:25}
\end{equation}
where the procedure $\Psi_n(\cdot)$ represents ResNet-50 for the $n$-th scale. Finally, the hybrid dual domain loss function can be formulated as
\begin{equation}
	\mathcal{L}_{hybrid} = \mathcal{L}_{mse} + \omega_2 \mathcal{L}_{msp}
	\label{eq:26}
\end{equation}
where $\omega_2$ is a weight parameter used to balance different loss components.
\subsection{Dataset}
Similar to previous MAR work (\citeauthor{9201079}, \citeyear{9201079}), we randomly choose 1200 FDCT images from the DeepLesion dataset (\citeauthor{8579063}, \citeyear{8579063}). Additionally, the metal masks are obtained from \cite{8331163}, encompassing 100 metal masks of varying location, shape, and size. Subsequently, we select 1000 FDCT images, and synthesize training dataset and validation dataset with 80 and 10 metal masks, respectively. The additional set of 10 metal masks is combined with an additional set of 200 CT images, generating a test dataset of 2000 CT images. The pixel sizes of the selected 10 metal masks for testing are: [2061, 890, 881, 451, 254, 124, 118, 112, 53, 35]. We simply categorize the first four masks as large metals and the latter six as small metals for performance evaluation.

For the selection of simulation parameters and scan geometry, we simulate the equiangular fan-beam projection geometry based on a 120 kVp polyenergetic X-ray source. We simulated three low-dose CT scenarios with Poisson noise in the sinogram. Specifically, we utilize incident X-ray containing $1\times10^{5}$ photons for 1/2 dose level, $5\times10^{4}$ photons for 1/4 dose level, and $2.5\times10^{4}$ photons for 1/8 dose level. We adjust the size of all images to $416\times416$ pixels and the size of the sinograms to $641\times640$ pixels.
\begin{table*}[t]
	\centering
	\caption{Quantitative evaluation of different approaches under 1/2 dose, under 1/4 dose and 1/8 dose levels. Best values are highlighted in bold. We report the PSNR (dB) $\textcolor{red}{\uparrow}$ / SSIM $\textcolor{red}{\uparrow}$ / RMSE (HU) $\textcolor{red}{\downarrow}$ values of the testing dataset for each case.}
	\renewcommand{\arraystretch}{1.2}
	{\footnotesize{\tabcolsep=4.5pt
			\begin{tabular}{c|c|c|c|c|c}
				\hline
				\multirow{1}{*}{\bf Algorithms}& \multicolumn{1}{c|}{\bf 1/2 Dose} & \multicolumn{1}{c|}{\bf 1/4 Dose}& \multicolumn{1}{c|}{\bf 1/8 Dose} &\multicolumn{1}{c|}{\bf Average} &\multicolumn{1}{c}{\bf  SD}\\ \hline
				\multicolumn{6}{c}{\bf Single-dose models} \\ 
				\hline
				\multicolumn{1}{c|}{FBP} & 27.81/0.4455/109.22&26.34/0.3523/127.68&23.99/0.2674/169.49&26.05/0.3551/135.46&1.57/0.0727/25.21\\
				\multicolumn{1}{c|}{LWFSN}&34.01/0.7177/53.80&32.83/0.6567/60.80&31.37/0.5973/71.16&32.74/0.6572/61.92&1.08/0.0492/7.13\\
				\multicolumn{1}{c|}{REDCNN}&38.76/0.9644/32.57&38.28/0.9601/34.07&37.71/0.9544/35.93&38.25/0.9596/34.19&0.43/0.0041/1.37\\
				\multicolumn{1}{c|}{CNN10}&37.44/0.9529/38.41&36.94/0.9448/40.32&36.42/0.9387/42.23&36.93/0.9455/40.32&{\bf 0.42}/0.0058/1.56\\
				\multicolumn{1}{c|}{EDCNN}&37.98/0.9558/35.96&37.54/0.9511/37.46&36.90/0.9447/39.92&37.47/0.9505/37.78&0.44/0.0045/1.63\\
				\multicolumn{1}{c|}{OctaveNet}&38.69/0.9635/32.85&38.12/0.9588/32.85&37.46/0.9521/37.12&38.09/0.9581/34.27&0.50/0.0047/2.01\\
				\multicolumn{1}{c|}{CTformer}&37.12/0.9435/39.62&36.57/0.9289/41.56&35.95/0.9241/44.15&36.55/0.9322/41.78&0.48/0.0083/1.86\\	
				\multicolumn{1}{c|}{LIT-Former}&37.34/0.9501/37.84
				&36.71/0.9436/40.34
				&35.98/0.9350/43.36
				&36.68/0.9429/40.51&0.56/0.0062/2.26\\
				\multicolumn{1}{c|}{InDuDoNet}&40.62/0.9651/27.19&40.03/0.9658/28.17&39.44/0.9598/29.78&40.03/0.9636/28.38&0.48/{\bf 0.0027/1.07}\\
				\multicolumn{1}{c|}{CLIPDenoising}&38.25/0.9619/34.13
				&37.70/0.9568/36.05
				&36.31/0.9051/41.42
				&37.42/0.9413/37.20&0.82/0.0257/3.09\\
				\multicolumn{1}{c|}{DuMSRNet}&\bf41.44/0.9750/22.91&\bf40.45/0.9712/26.41&\bf39.64/0.9649/28.06&\bf 40.51/0.9704/25.79& 0.74/0.0042/2.15\\
				\hline
				\multicolumn{6}{c}{\bf Muti-doses models}\\
				\hline
				\multicolumn{1}{c|}{AirNet}&36.10/0.9277/43.89&36.04/0.9344/43.97&35.46/0.9199/46.30&35.87/0.9273/44.72&0.29/0.0059/1.12\\
				\multicolumn{1}{c|}{PromptIR}
				&38.29/0.9630/36.55&37.99/0.9605/37.38&37.60/0.9573/38.55&37.96/0.9603/37.49&\textbf{0.28/0.0023/0.82}\\
				\multicolumn{1}{c|}{BDuMSRNet}&41.90/0.9766/22.45&41.18/0.9730/24.06&40.04/{\bf 0.9689}/27.15&41.04/{\bf0.9728}/24.54&0.77/0.0031/1.93 \\
				
				\multicolumn{1}{c|}{PDuMSRNet}&\bf 42.26/0.9771/20.76&\bf 41.58/0.9742/22.30&{\bf 40.44}/0.9667/{\bf 25.28}&{\bf 41.43}/0.9727/{\bf 22.78}&0.75/0.0044/1.88\\ \hline
	\end{tabular}}}
	\label{tab:1}
\end{table*}

\begin{table*} [!t]
	\setlength\tabcolsep{1pt}
	\centering
	\caption {Quantitative evaluation of different approaches under $1/8$ dose level, and their corresponding average and standard deviation values. Best values are highlighted in bold. We report PSNR (dB) $\textcolor{red}{\uparrow} $ / SSIM $\textcolor{red}{\uparrow} $ values of the testing dataset for each case.}
	\renewcommand{\arraystretch}{1.2}
	{\footnotesize{\tabcolsep=4.5pt
			\begin{tabular}{c|c|c|c|c|c|c|c}
				\hline
				\multirow{1}{*}{\bf Algorithms}&\multicolumn {2} {c} {\bf Large Metal} &\multicolumn {1} {c}{\bf ------------} &\multicolumn {2} {c|} {\bf Small Metal} &\multicolumn {1} {c|} {\bf Average} &\multicolumn {1} {c} {\bf SD}\\ 
				\cline {1-7} 
				\hline
				\multicolumn{8}{c}{\bf Single-dose models} \\ 
				\hline
				FBP
				&20.35/0.2433 &23.31/0.2638 &25.23/0.2742 &25.43/0.2757 &25.65/0.2802
				&23.99/0.2674 &2.00/0.0132
				\\
				LWFSN
				&29.61/0.5968 &30.63/0.5950 &31.99/0.5972 &32.23/0.5984 &32.39/0.5992 &31.37/0.5973&	\textbf{1.08/0.0014}
				\\
				REDCNN
				&33.98/0.9370 &35.83/0.9486 &38.79/0.9601 &39.72/0.9622 &40.23/0.9643 
				&37.71/0.9544 &2.41/0.0103
				\\
				CNN10
				&32.30/0.9126 &34.26/0.9294 &37.72/0.9473 &38.60/0.9505 &39.18/0.9537 
				&36.42/0.9387 &2.67/0.0155  
				\\
				EDCNN
				&32.84/0.9187 &34.75/0.9358 &38.14/0.9533 &39.09/0.9566 &39.67/0.9592&36.90/0.9447&2.65/0.0154
				\\
				OctaveNet
				&33.61/0.9339 &35.48/0.9460 &38.61/0.9579 &39.52/0.9602 &40.08/0.9622 &37.46/0.9520&	2.50/0.0107
				
				\\
				CTformer
				&32.18/0.9018 &34.03/0.9174 &37.17/0.9320 &37.93/0.9335 &38.42/0.9359&35.95/0.9241&	2.43/0.0129
				\\
				LIT-Former
				&32.62/0.9122 &34.34/0.9272 &37.05/0.9425 &37.76/0.9455 &38.14/0.9477
				&35.98/0.9350 &2.14/0.0135  
				\\ 
				InDuDoNet
				&33.62/0.9259 &{\bf 39.04}/0.9635 &\bf {41.32/0.9695} &\bf {41.61/0.9701} &\bf {41.59/0.9703}&39.44/0.9599&	3.06/0.0172
				\\
				CLIPDenoising
				&33.24/0.9029 &34.75/0.9027 &37.25/0.9059 &37.94/0.9065 &38.38/0.9077 
				&36.31/0.9051 &1.98/0.0020  
				\\
				DuMSRNet
				&\bf {35.89/0.9547} &38.84/\bf {0.9649} &40.78/0.9678 &41.28/0.9685 &41.40/0.9688 &\bf {39.64/0.9649} &2.17/0.0057  
				
				\\
				\hline
				\multicolumn{8}{c}{\bf Muti-doses models}\\
				\hline
				AirNet
				&31.88/0.9005 &33.69/0.9129 &36.64/0.9262 &37.35/0.9288 &37.76/0.9312
				&35.46/0.9199 &2.42/0.0125  
				\\
				PromptIR
				&32.42/0.9313 &34.67/0.9496 &39.07/0.9656 &40.44/0.9684 &41.37/0.9715
				&37.60/0.9573 &3.64/0.0154  
				\\
				BDuMSRNet
				&36.27/\bf {0.9587} &39.32/\bf {0.9691} &41.26/\bf {0.9717} &41.62/\bf {0.9723} &41.71/\bf {0.9728} &40.04/\bf {0.9689} &2.07/0.0053
				
				\\
				PDuMSRNet
				&\textbf{37.82}/0.9583 &\textbf{39.57}/0.9663&\textbf{41.39}/0.9692 &\textbf{41.66}/0.9696 &\textbf{41.75}/0.9700&\textbf{40.44}/0.9667&\bf 1.53/0.0044
				\\
				\bottomrule[1.2pt]
	\end{tabular}}}
	\label{tab:2}
\end{table*}

\subsection{Implementation Details and Evaluation Criteria}
We design three training strategies to evaluate performance of the proposed algorithm for LDMAR task, DuMSRNet, PDuMSRNet and blind DuMSRNet, i.e., BDuMSRNet. Specifically, DuMSRNet is trained at single-dose level to compare with other single-dose SOTA methods. In contrast, PDuMSRNet and BDuMSRNet are trained across all three dose levels. BDuMSRNet is the vision removing prompt guidance strategy compared with PDuMSRNet to valid the effect of prompt guidance strategy. During the training of PDuMSRNet, we utilize a total of 3000 images, comprising doses of 1/2, 1/4, and 1/8, with each dose containing 1000 images, to train a universal model. Conversely, during the training of DuMSRNet, we train three distinct models on images with 1/2, 1/4, and 1/8 dose levels, utilizing 1000 CT images for each dose level. Additionally, to further demonstrate the effectiveness of the prompt guidance strategy, similar to PDuMSRNet, we employ 3000 images to train DuMSRNet and obtain a single model, resulting in BDuMSRNet.

We implement our methods using the PyTorch framework, and the differentiable FP and FBP operations are implemented using the ODL library. In our experiments, we manually set $\alpha=0.5$, $\eta_1=1$, $\eta_2=5$, $\omega_1=0.01$, and $\omega_2=1\times10^{-4}$. The total number of epochs is set to 100, with a batch size of 1, utilizing an NVIDIA RTX 4090 GPU. We adopt the Adam optimizer with the parameters $(\beta_1,\beta_2)=(0.9,0.999)$ to optimize our network. The initial learning rate is set to $1\times10^{-4}$ and halved at epochs 40 and 80. The peak signal-to-noise ratio (PSNR), structured similarity index measure (SSIM), and root mean square error (RMSE) are employed for quantitatively evaluating the reconstruction outcomes.

\begin{figure*}[!t]
	\centering
	{\includegraphics[width=1.0\linewidth]{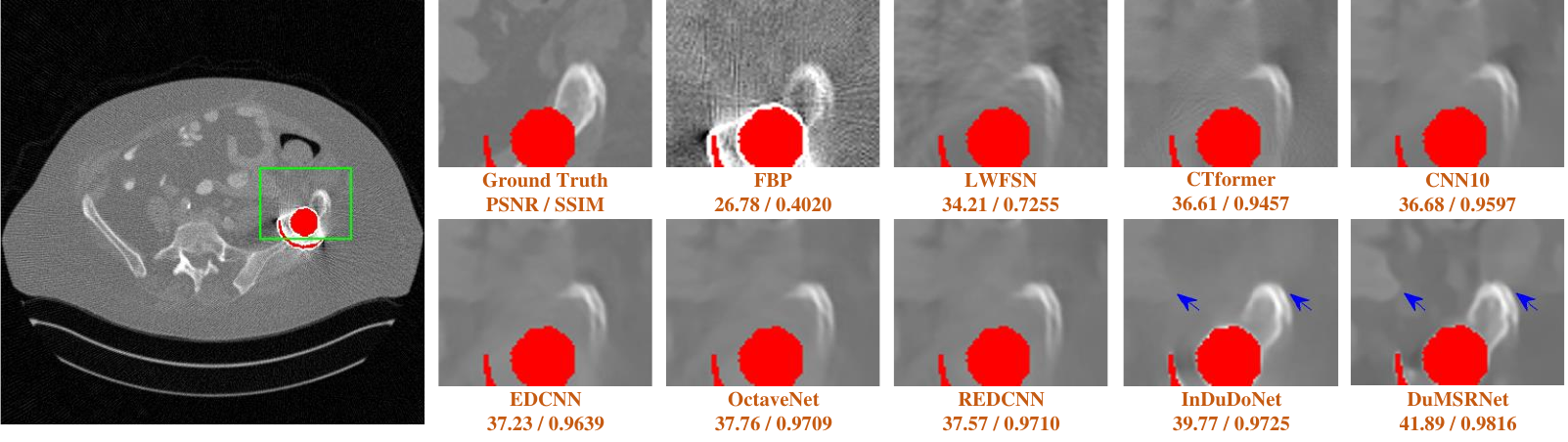}}
	\caption{{Visual evaluation of different approaches on LDCT images with metallic implants under the 1/2 dose level. Zooming in on the green screen enhances the viewing experience of the images. The PSNR and SSIM values are computed on the entire image reconstructed by the respective approaches. The metallic implants are highlighted with red masks. The blue arrow indicates the reconstructed structural information. The display window is [-1000, 1000] HU.}}
	\label{Fig:3}
	\vspace{-0.0cm}
\end{figure*}
\begin{figure*}[!t]
	\centering
	{\includegraphics[width=1.0\linewidth]{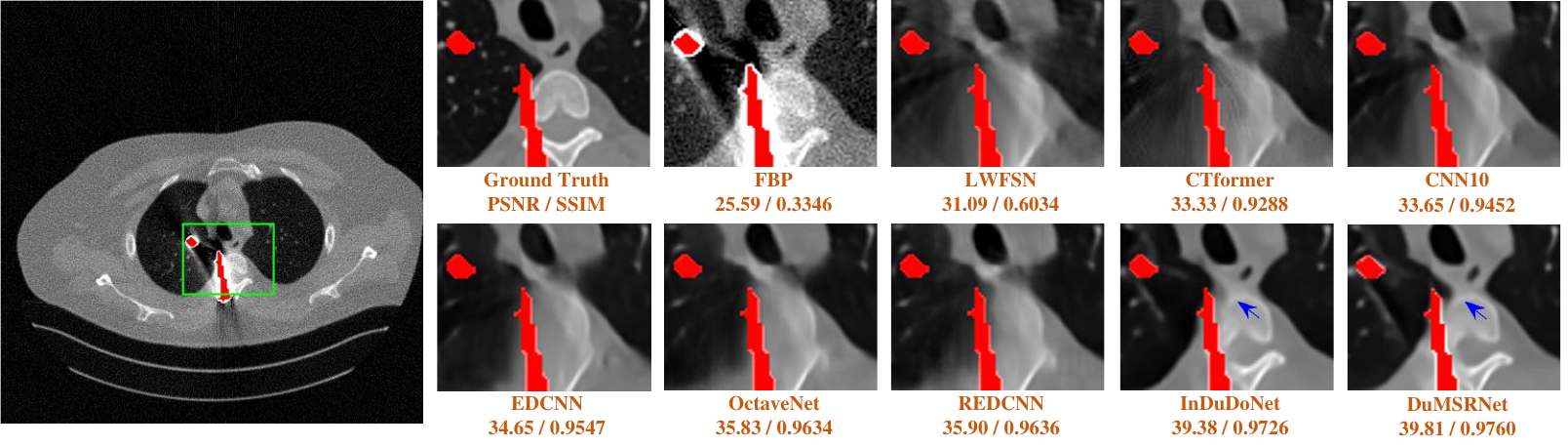}}
	\caption{{Visual evaluation of different approaches on LDCT images with metallic implants under the 1/4 dose level. Zooming in on the green screen enhances the viewing experience of the images. The PSNR and SSIM values are computed on the entire image reconstructed by the respective approaches. The metallic implants are highlighted with red masks. The blue arrow indicates the reconstructed structural information. The display window is [-1000, 1000] HU.}}
	\label{Fig:4}
\end{figure*}

\subsection{Performance Evaluation for LDMAR Task}
To demonstrate the superiority of DuMSRNet, we conduct both quantitative and qualitative comparisons with previous LDCT methods, such as FBP, LWFSN (\citeauthor{9721076}, \citeyear{9721076}), REDCNN (\citeauthor{7947200}, \citeyear{7947200}), CNN10 (\citeauthor{5401721}, \citeyear{5401721}), EDCNN (\citeauthor{9320928}, \citeyear{9320928}), OctaveNet (\citeauthor{10.1007/978-3-030-59354-4_7}, \citeyear{10.1007/978-3-030-59354-4_7}), CTformer (\citeauthor{Wang_2023}, \citeyear{Wang_2023}), LIT-Former (\citeauthor{Chen2023LITFormerLI}, \citeyear{Chen2023LITFormerLI}), InDuDoNet (\citeauthor{10.1007/978-3-030-87231-1_11}, \citeyear{10.1007/978-3-030-87231-1_11}), CLIPDenoising (\citeauthor{Cheng_2024_CVPR}, \citeyear{Cheng_2024_CVPR}), and multi-task learning approaches, such as AirNet (\citeauthor{airnet}, \citeyear{airnet}), PromptIR (\citeauthor{promptir}, \citeyear{promptir}). under different dose levels. All methods are trained and tested on the same dataset. Table~\ref{tab:1} reports the average PSNR, SSIM and RMSE values for the DeepLesion dataset at $1/2$, $1/4$ and $1/8$ dose levels. Table~\ref{tab:2} summarizes the average PSNR and SSIM values on varying metal sizes under $1/8$ dose level. As can be seen from Table~\ref{tab:1} and Table~\ref{tab:2}, all DL-based approaches outperform traditional methods in terms of PSNR and SSIM values, owing to the superior feature extraction and representation capabilities inherent to deep neural networks. Notably, the dual-domain frameworks, InDuDoNet and our DuMSRNet, achieve higher PSNR values compared to single-domain methods, underscoring the benefit of leveraging knowledge from both the sinogram and image domains. Our proposed PMDeepST demonstrates consistent and significant superiority across all evaluated dose levels and metal sizes. Fig.~\ref{Fig:3} and Fig.~\ref{Fig:4} illustrate the visual comparison between DuMSRNet and other SOTA methods under the 1/2 and 1/4 dose level and the visual comparisons critically illuminate the reasons behind our quantitative advantages. As observed, the FBP reconstructed images are severely degraded by noise and metal artifacts. While LWFSN, CNN10, REDCNN, CTformer, EDCNN, and OctaveNet achieve some artifact suppression, they concurrently introduce secondary noise and metal artifacts. InDuDoNet, though effective in removing most metal artifacts, tends to over-smooth the results, leading to a loss of fine anatomical details, which is associated with its ignoring on multi-scale feature within the CT image. By contrast, DuMSRNet more effectively suppresses a wider range of artifacts while simultaneously preserving critical tissue structures. This performance is directly linked to the core architectural strength of our method, the multi-scale adaptive sparse representation framework and the prompt guiding module, which is explicitly designed to model within-scale characteristics, cross-scale complementarity and the prior of metal implants. This design enables the network to distinguish between artifact patterns and true anatomical textures more reliably than methods relying on a single-scale or less adaptive representations. However, a critical analysis also reveals a limitation when processing larger metal implants. In these challenging cases, the network occasionally fails to distinguish massive streaking artifacts from true anatomical tissues, indicating that the current multi-scale sparsifying frames struggle to effectively model and remove large-scale metal artifacts. This observed shortcoming points toward a direction for future work aimed at enhancing the representational capacity for such extreme scenarios. \par
\begin{figure}[!t]
	\centering
	{\includegraphics[width=1.0\linewidth]{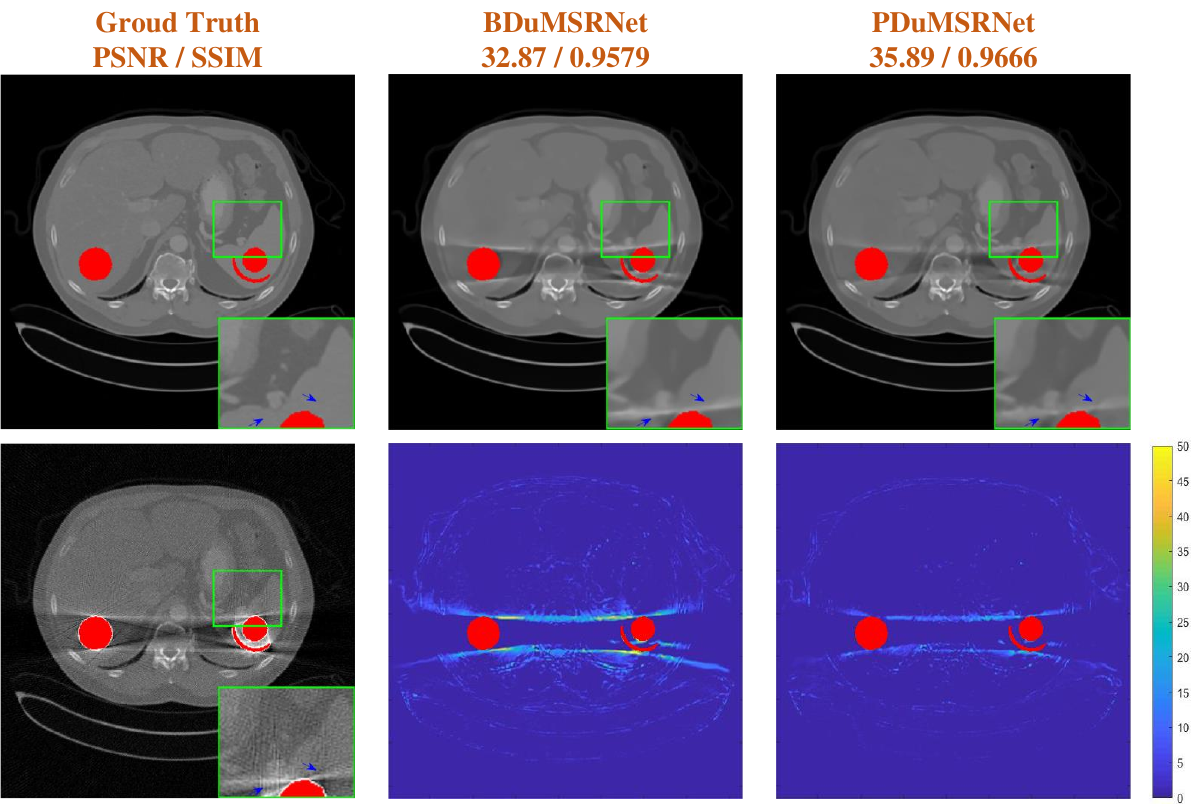}}
	\caption{{Visual evaluation and reconstruction error magnitude of BDuMSRNet and PDuMSRNet on LDCT images under the 1/4 dose level. Zooming in on the green screen enhances the viewing experience of the images. The PSNR and SSIM values are computed on the entire image reconstructed by the respective approaches. The metallic implants are highlighted with red masks. }}
	\label{Fig:5}
\end{figure}

To validate the efficacy of our prompt guiding strategy, we devised two models: the blind DuMSRNet, trained on a dataset comprising three doses without the prompt guiding strategy, and the PDuMSRNet, trained on a dataset with three doses using the prompt guiding strategy. As shown in Table~\ref{tab:1}, we observed a notable improvement of 0.39 dB in PSNR value for the PDuMSRNet compared to that of the Blind DuMSRNet. This substantial enhancement underscores the efficacy of the prompt guiding strategy. Leveraging prompt information enables the proposed DuMSRNet to better comprehend the intricate relationship between dose and noise, as well as the nuanced mapping between metal masks and artifacts, thereby achieving superior reconstructions. Fig.~\ref{Fig:5} illustrates the reconstruction outcomes obtained by BDuMSRNet and PDuMSRNet under the 1/4 dose level. As depicted, DuMSRNet exhibits superior noise and artifacts removal capabilities compared to BDuMSRNet, resulting in enhanced reconstruction performance. This improvement highlights the effectiveness of the prompt-embedding strategy. Moreover, since PDuMSRNet can achieve image reconstruction under three dose levels with a single model, it requires less model storage and offers greater flexibility than DuMSRNet. \par

\begin{table}[htbp]
	\centering
	\caption{Ablation studies of the dual domain learning under the 1/2 dose level. Best values are highlighted in bold.}
	\renewcommand{\arraystretch}{1.1}
	{\footnotesize{\tabcolsep=4pt\begin{tabular}{c|c|c|c} 
				\hline
				{\bf Network} &{\bf PSNR (dB) $\textcolor{red}{\uparrow}$} & {\bf SSIM $\textcolor{red}{\uparrow}$} & {\bf RMSE (HU) $\textcolor{red}{\downarrow}$} \\ 
				\hline
				Sinogram-Domain-Only &34.52 &0.8545&51.19 \\
				Image-Domain-Only &38.30 &0.9624 &33.78 \\
				\hline
				DuMSRNet &\bf 41.44 &\bf 0.9750 &\bf 22.91 \\
				\hline
	\end{tabular}}}
	\label{tab:3}
\end{table}
\subsection{Ablation Studies}
{\bf (1) The impact of dual domain learning:}
During each unfolding stage, DuMSRNet includes a network for the sinogram domain and a network for the image domain. To showcase the effect of dual domain learning, we train sinogram-domain-only and image-domain-only networks under identical conditions. We then compare their performance with DuMSRNet using PSNR, SSIM, and RMSE metrics. As summarized in Table~\ref{tab:3}, it is evident that our dual domain approach (i.e., DuMSRNet) outperforms the sinogram-domain-only and image-domain-only methods, demonstrating the effectiveness of leveraging joint dual domain information. \par 
\begin{table}[htbp]
	\centering
	\caption{Ablation studies on different modules of DuMSRNet under the 1/2 dose level. We report the PSNR (dB) / SSIM / RMSE (HU) values of the testing dataset for each case. Best values are highlighted in bold.}
	\renewcommand{\arraystretch}{1.1}
	{\footnotesize{\tabcolsep=6pt\begin{tabular}{c|c|c|c} 
				\hline
				{\bf Network} &{\bf PSNR $\textcolor{red}{\uparrow}$} & {\bf SSIM $\textcolor{red}{\uparrow}$} & {\bf RMSE $\textcolor{red}{\downarrow}$} \\ 
				\hline
				 w/o SATG &40.65 &0.9725 &22.91 \\
				 w/o LIEB &41.09 &0.9735 &23.89 \\
				 w/o RIEB &41.18 &0.9736 &23.53 \\
				 w/o GIEB &40.96 &0.9718 &24.34 \\
				 w/o FEM &41.16 &0.9735 &23.85 \\
				baseline &\bf 41.44 &\bf 0.9750 &\bf 22.91 \\
				\hline
	\end{tabular}}}
	\label{tab:4}
\end{table}

{\bf (2) The impact of the scale-adaptive threshold generator:}
Considering the crucial role of thresholds during the shrinking procedure, we construct a prompt guiding scale-adaptive threshold generator. To validate the effectiveness of prompt guiding scale-adaptive threshold generator (PSATG), we designed two groups of experiments. The first group involves DuMSRNet models at a specific dose of 1/2 to validate the effectiveness of the scale-adaptive threshold generator. \par
We employ a learnable parameter as the threshold, replacing the threshold generated by SATG, forming DuMSRNet w/o SATG. Note that SATG here does not contain the prompt guiding module. When combined with DuMSRNet and DuMSRNet w/o SATG, the proposed SATG produces a significant performance improvement on all metrics, as summarized in Table~\ref{tab:4}. To assess the effectiveness of the combination of the deep feature extraction module with three branches and the feature selection module in SATG, we selectively remove the corresponding parts during training, resulting in DuMSRNet w/o LIEB, DuMSRNet w/o RIEB, DuMSRNet w/o GIEB, and DuMSRNet w/o FEM. The corresponding quantitative results are summarized in Table~\ref{tab:4}. The comparison among DuMSRNet w/o LIEB, DuMSRNet w/o RIEB, DuMSRNet w/o GIEB, and DuMSRNet validates that the combination of local, regional, and global information significantly enhances the reconstruction performance. Additionally, compared to DuMSRNet w/o FEM, DuMSRNet achieves better performance, thanks to the capability of the feature selection module. Specifically, the feature selection module can reweight local, regional, and global information, preventing performance degradation caused by information redundancy.\par
\begin{table}[htbp]
	\centering
	\caption{Ablation studies on different prompts of PDuMSRNet under the 1/2, 1/4 and 1/8 dose levels. We report the PSNR (dB) / SSIM / RMSE (HU) values of the testing dataset for each case. Best values are highlighted in bold.}
	\renewcommand{\arraystretch}{1.1}
	{\footnotesize{\tabcolsep=4pt\begin{tabular}{c|c|c|c} 
				\hline
				{\bf Prompt} &{\bf PSNR $\textcolor{red}{\uparrow}$} & {\bf SSIM $\textcolor{red}{\uparrow}$} & {\bf RMSE $\textcolor{red}{\downarrow}$} \\ 
				\hline

				input+mask &34.97 &0.9290 &69.29\\
				input+1/dose & 35.40&0.9479 &61.30 \\
			1/dose+mask &37.88&0.9605 &36.50 \\
				input+dose+mask &33.75&0.9280 & 91.05 \\
				baseline &\bf 41.43&\bf 0.9727&\bf 22.78\\
				\hline
	\end{tabular}}}
	\label{tab:5}
\end{table}
{\bf (3) The impact of prompt guiding strategy:}
The second group pertains to PDuMSRNet models across multiple dose levels (1/2, 1/4, and 1/8) to validate the effectiveness of the prompt guiding strategy. To facilitate the adaptation of a single model to various CT dose settings, we introduce a prompt guiding strategy that incorporates the prompt information from input LDCT images with metallic implants, mental masks, and dose maps to guide the training process. To evaluate the effectiveness of different prompts, we systematically remove input LDCT images with metallic implants, mental masks, and dose maps, and replace the 1/dose map with the dose map in separate instances. As illustrated in Table~\ref{tab:5}, the performance in these ablation experiments falls short of that of the original PDuMSRNet, thereby confirming the validity of each individual prompt.\par

\begin{table}[htbp]
	\centering
	\caption{Ablation studies on MSFuM of DuMSRNet under the 1/2 dose level. We report the PSNR (dB) / SSIM / RMSE (HU) values of the testing dataset for each case. Best values are highlighted in bold.}
	\renewcommand{\arraystretch}{1.1}
	{\footnotesize{\tabcolsep=6pt\begin{tabular}{c|c|c|c} 
				\hline
				{\bf Network} &{\bf PSNR $\textcolor{red}{\uparrow}$} & {\bf SSIM $\textcolor{red}{\uparrow}$} & {\bf RMSE $\textcolor{red}{\downarrow}$} \\ 
				\hline

			 w/o MSFuM &40.55 &0.9664 &25.19 \\
				with MSFuM &\bf 41.44 &\bf 0.9750 &\bf 22.91 \\
				\hline
	\end{tabular}}}
	\label{tab:6}
\end{table}

{\bf (4) The impact of fusing high-resolution and low-resolution information:}
To fully utilize cross-scale complementarity, we design a multi-scale coefficient fusion module using the cross attention mechanism, achieving efficient fusion of multi-scale information. To validate the effectiveness of fusing high-resolution and low-resolution information, we modify the original network architecture to create DuMSRNet w/o MSFuM, where the architecture no longer performs the fusion of high-resolution and low-resolution information. As shown in Table~\ref{tab:6}, the fusion mechanism of high-resolution and low-resolution leads to an improvement in PSNR values by 0.89 dB, SSIM values by 0.0086, and a reduction in RMSE values by 2.28.\par

\begin{figure*} [!t]
	\centering
	{\includegraphics[width=1.0\linewidth] {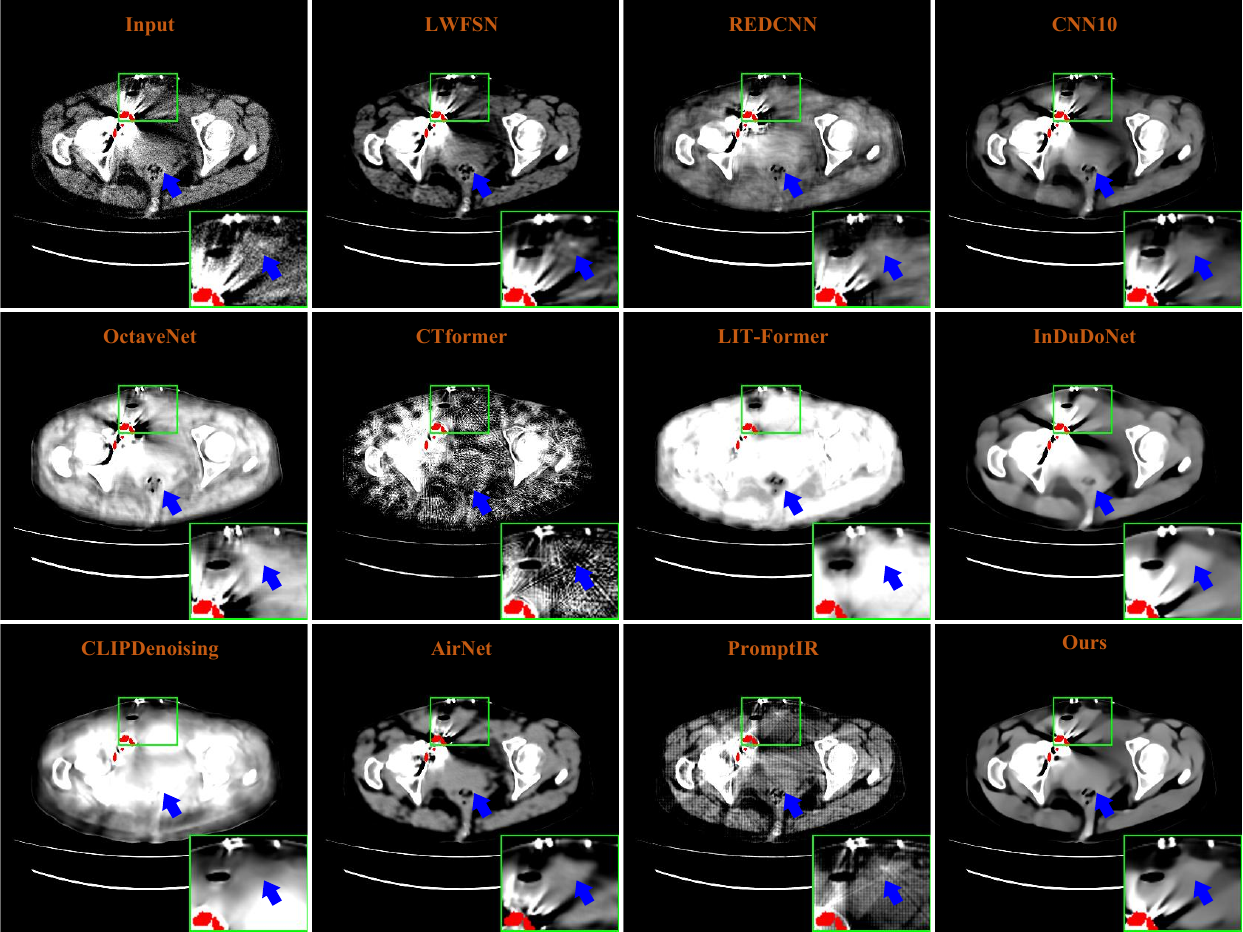}}
	\caption {Visual evaluation of different approaches on clinical dataset with metallic implants. The magnified details are shown in green box. The display window is [-175, 275] HU. Blue arrows indicate key areas, such as tissues near the metallic artifacts, which highlight differences in reconstruction quality among all methods.}
	\label {Fig:6}
	
\end{figure*} 
\begin{figure}[!t]
	\centering
	{\includegraphics[width=1\linewidth]
		{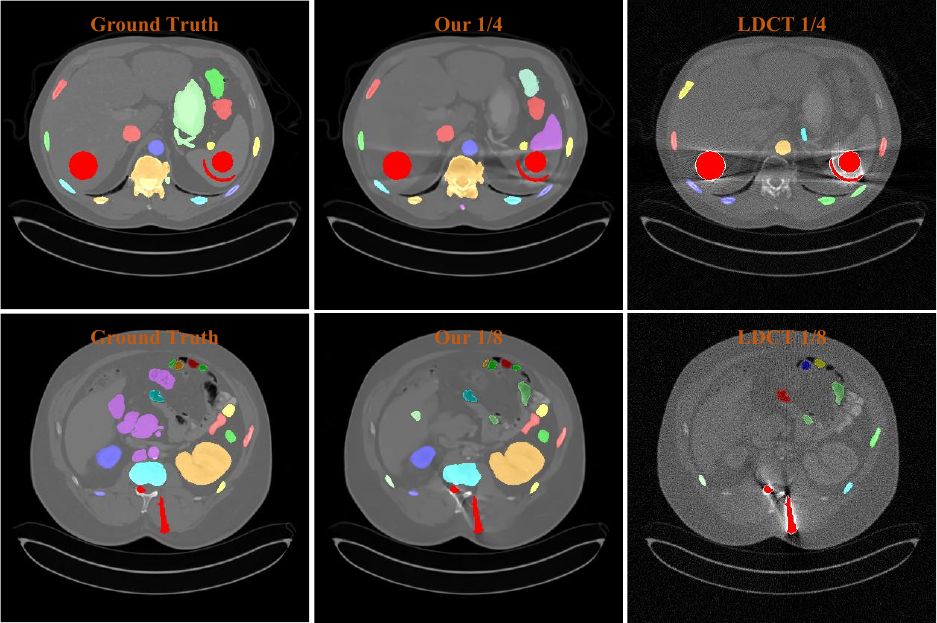}}
	\caption {SAM segmentation results for LDCT images and results of our proposed method. Different colors represent distinct segmented regions. Blue arrows indicate key areas, such as tissues near the metallic implants and artifacts, which highlight differences in segmentation quality.}
	\label {Fig:7}
\end{figure}
\section{Discussion}
\subsection{Analysis on Storage Requirement}
To demonstrate the superior storage efficiency of DuMSRNet, we conduct a qualitative comparison of model storage requirements with the traditional single-dose models (DuMSRNet) and the multi-dose models (BDuMSRNet and PDuMSRNet). As illustrated in Table~\ref{tab:7}, DuMSRNet incurs the highest storage costs among the models. The storage requirement of PDuMSRNet is approximately equivalent to BDuMSRNet, which is one third of DuMSRNet trained model for each dose level. Although the storage cost of the PDuMSRNet is slightly higher than that of the BDuMSRNet, the reconstruction quality achieved by the PDuMSRNet is better than that achieved by the BDuMSRNet. This performance gain is attributable to the inherent capacity of the multi-task, multi-setting architecture of PDuMSRNet, which leverages shared representations while adapting to specific task conditions through its prompt-guided modules.
\begin{table}[htbp]
	\centering
	\caption{Storage requirement for different models. In comparison regarding the storage costs (KB) of models for three doses, the multiple-in-one model demonstrates better storage efficiency}
	\renewcommand{\arraystretch}{1.1}
	{\footnotesize{\tabcolsep=6pt\begin{tabular}{c|c|c|c}
				\hline
				{\bf} &{\bf Single-dose models} &\multicolumn{2}{|c}{\bf Muti-doses models} \\ 
				\hline
				{\bf Algorithms} &{\bf DuMSRNet} & {\bf BDuMSRNet} & {\bf PDuMSRNet} \\ 
				\hline
				Storage &158,856 &52,952 &52,996 \\
				\hline
	\end{tabular}}}
	\label{tab:7}
\end{table}
\subsection {Analysis on Clinical Validation}
To further validate the clinical applicability of our method, we conducted validation on the clinical dataset, obtained from the public CTPelvic1K library \footnote{CTPelvic1K: \url{https://github.com/MIRACLE-Center/CTPelvic1K}}. CTPelvic1K dataset is a publicly accessible resource specifically designed for pelvic fracture segmentation. The dataset includes 14 testing volumes labeled with multi-bone, i.e., sacrum, left hip, right hip, and lumbar spine. We evaluate the clinical performance of our model on the CTPelvic1K dataset, employing weights pre-trained on the simulated DeepLesion dataset to assess its generalization capability. Fig.~\ref {Fig:6} presents visualization evaluation of different approaches on clinical dataset with metallic implants. \par 	
As illustrated in Fig.~\ref{Fig:6}, due to low dose and the presence of metallic implants, significant noise and artifacts are observed around the metal regions and algorithms exhibit different reconstruction performance from simulated dataset. Specifically, methods such as OctaveNet, LIT-former and CLIPDenoising exhibit severe distortion and highlight clipping in their reconstruction results, significantly harming diagnostic utility. While LWFSN, REDCNN, CNN10 and InDuDoNet demonstrate more structural coherence, they fail to adequately suppress metal artifacts. CTformer and PromptIR introduce secondary textural artifact while they achieve the most effective metal artifacts reduction. Although AirNet demonstrates satisfactory performance in metal artifact reduction and denoising, its output exhibits significant blurring in critical edge regions, as indicated by the blue arrows in Fig.~\ref{Fig:6}. This degradation of edge detail represents a notable limitation for applications requiring precise structural delineation. In contrast, the proposed PDuMSRNet effectively addresses this weakness. It achieves optimal performance by successfully preserving crucial edge information while simultaneously maintaining high efficacy in both artifact suppression and noise reduction.\par
\subsection {Analysis on Segmentation Tasks}
To demonstrate the clinical significance of our proposed method PDuMSRNet, we evaluated its impact on downstream segmentation tasks. Specifically, the Segment Anything Model (SAM) (\citeauthor{sam}, \citeyear{sam}), a SOTA segmentation framework, was deployed using its official implementation. Both original LDCT images with metallic implants and our reconstructed results were processed by the identical SAM instance. As illustrated in Fig.~\ref{Fig:7}, our reconstruction results enabled the segmentation of significantly more discernible anatomical structures compared to the LDCT inputs with metallic implants. This quantitative improvement in segmentable regions directly enhances the clinical utility of the images, validating efficacy of our method in facilitating critical downstream applications such as diagnostic analysis and treatment planning.

\section{Conclusion}
In this paper, we proposed a model interpretable prompt guiding multi-scale adaptive sparse representation-driven network named PMSRNet for LDMAR, and further constructed PDuMSRNet, an enhancement of DuMSRNet in dual domain.  Specifically, the PMSRNet consists of three components: multi-scale sparsifying frames, PSATG and MSFuM. The proposed multi-scale sparsifying frames could simultaneously employ within-scale characteristics and cross-scale complementarity. The elaborated PSATG could extract features at the local, regional, and global range to capture multiple contextual information. Additionally, the introduced MSFuM could fuse detailed information from high-resolution features and contextual information from low-resolution features. Extensive experiments demonstrated that the proposed DuMSRNet consistently outperformed the state-of-the-art methods and the proposed PDuMSRNet achieved less model storage cost and more flexibility. However, current methods exhibit blurring in clinical dataset. This limitation is fundamentally attributable to insufficient model generalization capability and the inherent domain gap between simulated training data and clinical test data. Future work will involve training the model using real-world clinical datasets to enhance reconstruction performance. 
\section*{Acknowledgments}
This work was supported by the National Natural Science Foundation of China under Grants 62371414, by the Natural Science Foundation of Hebei Province under Grant F2025203070, and by the Hebei Key Laboratory Project under Grant 202250701010046.
\bibliographystyle{model2-names}
\biboptions{authoryear}
\bibliography{refs}

\end{document}